\documentclass[11pt, a4paper, logo, copyright, nonumbering]{deepseek}
\usepackage[numbers, square, sort&compress]{natbib}
\usepackage{dblfloatfix}
\usepackage{ulem}
\usepackage{caption}
\definecolor{citecolor}{HTML}{1976D2}
\hypersetup{
    colorlinks=true,
    linkcolor=black,
    filecolor=magenta,
    urlcolor=cyan,
    citecolor=citecolor,
}
\usepackage{dramatist}
\usepackage{xspace}
\usepackage{pifont}
\usepackage{multirow}
\usepackage{tcolorbox}
\usepackage{xltabular}
\usepackage{longtable}
\usepackage{hyperref}
\usepackage{diagbox}
\usepackage{makecell}
\usepackage{amsfonts}
\usepackage{amsmath}
\usepackage{amssymb}
\usepackage{lineno}

\usepackage[bottom]{footmisc}

\usepackage{CJKutf8}
\usepackage{subfigure}
\usepackage{setspace}
\usepackage{subcaption} 
\usepackage{epigraph}
\usepackage{wrapfig}

\makeatletter
\def\@BTrule[#1]{%
  \ifx\longtable\undefined
    \let\@BTswitch\@BTnormal
  \else\ifx\hline\LT@hline
    \nobreak
    \let\@BTswitch\@BLTrule
  \else
     \let\@BTswitch\@BTnormal
  \fi\fi
  \global\@thisrulewidth=#1\relax
  \ifnum\@thisruleclass=\tw@\vskip\@aboverulesep\else
  \ifnum\@lastruleclass=\z@\vskip\@aboverulesep\else
  \ifnum\@lastruleclass=\@ne\vskip\doublerulesep\fi\fi\fi
  \@BTswitch}
\makeatother

\addto\extrasenglish{
}

 {\begin{list}{}%
         {\setlength{\leftmargin}{#1}}%
         \item[]%
 }
 {\end{list}}
 
\reportnumber{001} 

\renewcommand{\today}{}

\title{\centering
    Unlimited OCR Works\\
    {\large\normalfont\textit{Welcome the Era of One-shot Long-horizon Parsing}}
}

\author{
 Baidu Inc.\\
}

\renewcommand{\phi}{\varphi}

\renewcommand{\epsilon}{\varepsilon}
\renewcommand{\imath}{\mathrm{i}}

\newlength{\restsubwidth}
\newlength{\restsubheight}
\newlength{\restsubmoreheight}
\newcommand{\rest}[2]{%
        \settowidth{\restsubwidth}{\ensuremath{#2}}
        \settoheight{\restsubheight}{\ensuremath{{}_{#2}}}
        \ensuremath{{#1\hskip 0.5pt}_{\vrule\kern2pt\parbox[b][%
        4pt][b]{\the\restsubwidth}{%
                        \ensuremath{{}_{#2}}}}}
        }

\begin{abstract}
Recently, end-to-end OCR models, exemplified by DeepSeek OCR, have once again thrust OCR into the spotlight. A widely held view is that employing a large language model (LLM) as the decoder allows the model to leverage the prior distribution of language, leading to improved OCR performance. However, the downside is equally evident: as the output sequence lengthens, the accumulated KV cache drives up memory consumption and progressively slows down generation. This stands in stark contrast to humans, who exhibit no such decline in efficiency during long-horizon copying tasks. In this technical report, we propose Unlimited OCR, a model designed to emulate human parsing working memory. Taking DeepSeek OCR as the baseline, we replace all attention layers in the decoder with our proposed Reference Sliding Window Attention (R-SWA), which reduces attention computation costs while maintaining a constant KV cache throughout the entire decoding process. By combining the high compression rate of DeepSeek OCR's encoder with our constant KV cache design, Unlimited OCR can transcribe dozens of pages of documents in a single forward pass under a standard maximum length of 32K. More importantly, R-SWA is a general-purpose parsing attention mechanism — beyond OCR, it is equally applicable to tasks such as ASR, translation, etc. Codes and model weights are publicly available at \url{http://github.com/baidu/Unlimited-OCR}.

\end{abstract}

\begin{document}
\begin{CJK*}{UTF8}{gbsn}

\maketitle

\begin{figure}[h]
    \centering
    \includegraphics[width=0.88\linewidth]{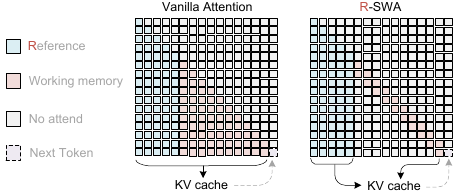}
    \caption{Illustration of Reference Sliding Window Attention (R-SWA). Each generated token attends to all reference tokens (visual tokens in OCR) and the preceding $n$ output tokens (128 by default). Compared to standard full attention, R-SWA maintains a constant KV cache throughout decoding. Compared to vanilla SWA, it preserves visual token fidelity by excluding them from state transitions, thereby avoiding progressive blurring.}
    \label{fig:1}
\end{figure}

\newpage

\begin{spacing}{0.9}
\tableofcontents
\end{spacing}

\newpage

\section{Introduction}

Humans are remarkably adept at seemingly straightforward long-horizon tasks: transcribing hundreds of book pages, translating hours-long audio recordings, and the like. Yet these are precisely the tasks where current models fall short. Take OCR as an example—no existing model~\cite{wei2025deepseek,wei2024general,cui2025paddleocrvl,wang2024mineru} can even parse ten of pages in a single forward pass. Instead, they resort to page-by-page processing in a for-loop fashion, resetting memory at every step. This divergence is far from superficial, and it cannot be reduced to a mere lack of sufficient context. When humans perform such tasks, they maintain a continuous cognitive state in which distant outputs fade softly from memory, while nearby context is used to track progress. The for-loop paradigm, by contrast, erases memory entirely at each page, fragmenting a coherent long-horizon process into isolated short tasks managed by an external scheduler. It works to some extent, but it remains an engineering workaround, not a step toward AGI-purpose intelligence.

Consider the act of transcribing a document. As we copy each character, we do not scan the entire text already written; we simply glance at the immediately surrounding context to stay oriented. This everyday behavior points to an attention pattern fundamentally different from those in current models. It is not standard full attention—the full history is never fully consulted. Nor does it resemble linear attention, since visual/reference tokens undergo no recurrent state updates; such updates would progressively blur the visual features and degrade recognition accuracy. To align more closely with this natural attention flow, and to explore how multimodal large language models (MLLMs)~\cite{team2023gemini,Qwen2.5-VL,huang2026step3,GPT4} can handle simple long-horizon parsing tasks, we propose Unlimited OCR. Our main contributions are as follows:

\begin{itemize}
\item We introduce Reference Sliding Window Attention (R-SWA), illustrated in Figure~\ref{fig:1}. For each token, R-SWA attends to all reference tokens—visual tokens and the prompt—while limiting output attention to the preceding $n$ tokens ($n$ defaults to 128). In this way, each token perceives the full image and autonomously tracks OCR progress through state transitions within the causal sliding window. This design keeps the KV cache constant during inference, alleviating memory pressure and reducing the computational cost. 


\item Building on R-SWA, we propose Unlimited OCR. Using DeepSeek OCR as our baseline, we retain its DeepEncoder with high image compression rate, modifying all the decoder LLM's attention mechanism to R-SWA. This enables Unlimited OCR to parse dozens of paper pages in a single forward pass. R-SWA also yields a modest improvement in general OCR accuracy. Specifically, Unlimited OCR achieves 93\% on the OmniDocBench v1.5 benchmark~\cite{ouyang2025omnidocbench}, outperforming the DeepSeek OCR baseline by 6\%.

\item We conduct a preliminary validation of MLLM architectures with linear-complexity attention on OCR tasks, particularly in long-horizon scenarios. Rather than brute-force scaling up the training context, we identify an elegant approach that achieves long-horizon OCR. Looking ahead, we see promise in extending R-SWA to ASR, translation, and other reference-based tasks that demand long-horizon dependency modeling.
\end{itemize}

In summary, we present R-SWA, which substantially reduces both the computational cost of attention and the memory footprint in the long-horizon inference. Building on R-SWA, Unlimited OCR not only enables one-shot parsing of an entire book, but also surpasses the DeepSeek OCR baseline by a large margin on popular document parsing benchmarks. Furthermore, we believe R-SWA holds promise well beyond OCR.


\section{Related Works}
\label{sec:related1}


\subsection{Pipeline-based Framework}
Traditional OCR models, particularly those designed for document parsing, typically adopt a pipeline architecture~\cite{cui2025paddleocr,cui2025paddleocrvl,wang2024mineru,li2025monkeyocr,feng2025dolphin}: a detection model first identifies different types of document elements, followed by multiple recognition operators that further parse the content within those blocks. These components are often bridged by a variety of heuristic strategies, such as cropping, rectification, and so on. In recent years, with the powerful decoder capabilities of large language models (LLMs), the pipeline-based OCR paradigm has continued to evolve~\cite{li2025monkeyocr}. The most straightforward adaptation retains the detection model while consolidating the multiple recognition models into a single unified model---a pragmatic hybrid that combines mature traditional detection algorithms with the advanced decoder of an LLM. Beyond this, there is another pipeline variant that invokes the LLM twice, replacing even the detection model with the same LLM~\cite{feng2025dolphin}, so that the entire OCR workflow becomes: LLM detection--cropping strategy--LLM recognition. Thanks to the inherent flexibility in how OCR tasks can be decomposed, pipeline architectures still remain widely adopted to this day.

\subsection{End-to-end Model}
With the advancement of vision-language models (VLMs)~\cite{li2023blip,Qwen-VL,Qwen2.5-VL,wei2024vary,huang2026step3} , end-to-end OCR, especially dense OCR models~\cite{blecher2023nougat,wei2024general,wei2025deepseek,wei2026deepseek,dots,poznanski2025olmocr} are on the rise. This approach fully leverages the powerful decoder capabilities of LLMs by merging text detection and recognition into a single unified function, allowing the entire content of a page to be parsed in a single forward pass. Compared with the pipeline approach, the end-to-end algorithm places higher demands on model capacity and poses greater training challenges. This, in turn, makes research on end-to-end OCR models all the more compelling: innovations in architectural design and iterative improvements in training methodologies can more directly inspire, or even advance, the development of general-purpose VLMs. 


\subsubsection{High-compression Encoder}
In end-to-end models, the encoder is an indispensable module that extracts and compresses image information. To a certain extent, the encoder determines the upper bound of the model: taking generation efficiency as an example, if the input vision tokens are too long—meaning the encoder's token compression ratio is insufficient—the model's decoding efficiency will be hindered by excessively long prefix tokens, thereby affecting decoding speed. The same holds true for effective decoding length. DeepEncoder~\cite{wei2025deepseek} achieves a 16$\times$ token compression rate under low activation values by cascading window attention ViT~\cite{kirillov2023segment} and global attention one~\cite{radford2021learning}, making it an ideal choice for multi-page long-horizon OCR. 

\subsubsection{High-efficiency Decoder}
What most directly affects inference cost is the decoder, including the activation value of the LLM and the KV cache size. Regarding the former, current end-to-end OCR models are typically under 3B parameters. In a related vein, DeepSeek OCR~\cite{wei2025deepseek} uses an MoE architecture~\cite{deepseek32}, keeping its activation at only 500M during inference. As for the KV cache, current models all see it grow continuously with decoding contexts, which limits both generation speed and length. This is exactly the key issue that our Unlimited OCR aims to address.

\begin{figure}[ht]
	\centering
    \includegraphics[width=1.0\linewidth]{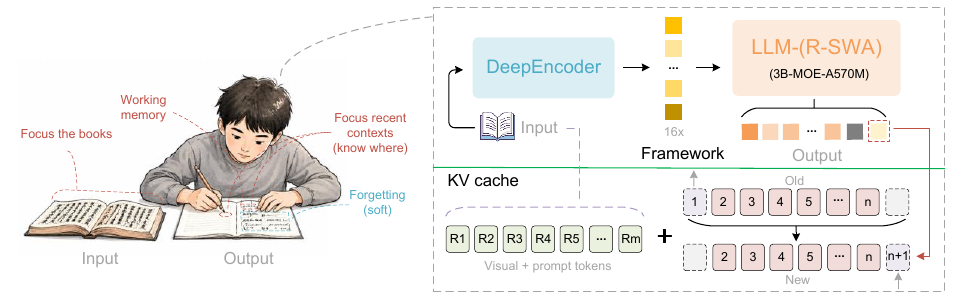}
	\caption{Inspired by the process of humans copying books, we propose the Unlimited OCR. This model features a unified end-to-end architecture, consisting of an encoder and a MoE-LLM decoder in which all attention mechanisms are R-SWA. The KV cache is implemented as a queue with a capacity of $m+n$—each time a new token is generated, the KV corresponding to the $(m+1)$-th token in the queue is evicted, ensuring that both computational cost and memory usage do not progressively increase during the generation process.}
	\label{fig:architecture}
\end{figure}

\section{Methodology}

\subsection{Long-horizon Parsing}
Our humans excel at long-horizon parsing tasks—continuously transcribing an entire book, translating even hundreds of pages in one sitting, or transcribing hours of audio without interruption. This continuous parsing capability appears closely linked to the working memory. As illustrated in Figure~\ref{fig:architecture}, when a person copies a book by hand, their attention typically centers on three points: the original source book, a small portion of what has just been written (usually only a few characters), and the next character about to be written. Rather than retaining a complete memory of everything already transcribed, they engage in a form of soft forgetting. This maybe the key to sustaining long-horizon parsing under low cognitive load. Inspired by this observation, we present Unlimited OCR.

\subsection{Architecture}

As shown in Figure~\ref{fig:architecture},  Unlimited OCR adopts DeepSeek OCR as its baseline. Specifically, it comprises the DeepEncoder paired with a Mixture-of-Experts (MoE) architecture that enjoys 3B total and 500M activated parameters. The DeepEncoder stands out for its exceptional visual token compression capability, which can dramatically reduce the KV cache footprint during the prefill stage while preserving robust optical text feature extraction. Departing from the original DeepSeek OCR, we replace the vanilla Multi-Head Attention (MHA) with our proposed R-SWA. With the new proposed attention, long-horizon parsing can be achieved by augmenting the original reference KV cache $m$ with a fixed-capacity output KV buffer of width $n$. We will delve into the technical details in the following sections.

\subsection{DeepEncoder}
DeepEncoder is originally introduced in DeepSeek OCR~\cite{wei2025deepseek}. It cascades SAM-ViT~\cite{kirillov2023segment} with CLIP-ViT~\cite{radford2021learning} and applies 16$\times$~\cite{wei2024vary} token compression at the bridge, so that the first half relies entirely on window attention to process the original image tokens, while global attention is reserved exclusively for the compressed tokens. This design keeps the activation values low when encoding high-resolution images, thereby conserving GPU memory. DeepEncoder natively supports five resolution modes; we retain two of them:  the "Base" model (1024×1024 for multi-page), and the "Gundam" mode (dynamic resolution for single-page). Specifically, DeepEncoder can compress a 1024$\times$1024 PDF-image to just 256 tokens. This high compression ratio is critically important for unlimited OCR works, because visual tokens do not undergo state transitions alongside the output - they are encoded once and remain static throughout the entire long-horizon parsing process.

\subsection{Reference Sliding Window Attention}

Despite the satisfactory compression of visual tokens that DeepEncoder achieves on the input side, the real bottleneck for one-shot parsing of an entire book lies in the decoding stage. Assume a compression ratio of 1:10 between visual and text tokens — \textit{i.e.}, one visual token can decode around ten text tokens. In that case, 10K visual tokens (equivalent to roughly $20-30$ pages at 1024$\times$1024 resolution) demand an output length of 100k$+$ tokens for full decoding. This has long been a formidable challenge for vanilla LLM-driven OCR models, due to the massive KV cache storage and attention computation that sequences beyond 128k tokens entail. To address this, we propose Reference Sliding Window Attention (R-SWA).

\subsubsection{Attention computation}

In essence, R-SWA constrains attention within a two-segment window of size $m+n$, as illustrated in Figure~\ref{fig:architecture}. Here, $m$ denotes the window for prefix tokens, which includes both visual tokens and the prompt. During a single inference pass, 
$m$ remains fixed; it depends only on the number of book pages or the resolution size of the document being decoded, and does not vary with decoding length. The window $n$ for the decode region is also fixed in size and slides in a causal manner. Specifically, the formulation is as follows:
\begin{align}
\mathcal{N}(t) &= \mathcal{P}\cup \mathcal{D}_n(t); \ \ \ \ \mathcal{P} = \{1,\dots,L_m\}, \\
\mathcal{D}_n(t) &= \left\{\,j \mid \max(L_m+1,\;L_m+t-n)\le j \le L_m+t-1 \right\},
\end{align}
where $\mathcal{P}$ denotes the prefix segment of length $L_m$, which is globally visible to all subsequent tokens, and $\mathcal{D}_n(t)$ denotes the causal sliding window of width $n$ over the decode region. The attention weight from token $t$ to position $j\in\mathcal{N}(t)$ is then computed as
\begin{align}
\alpha_{tj}
&=
\frac{
\exp\left(\frac{\mathbf{q}_t^\top \mathbf{k}_j}{\sqrt{d_k}}\right)
}{
\sum\limits_{i\in\mathcal{N}(t)}
\exp\left(\frac{\mathbf{q}_t^\top \mathbf{k}_i}{\sqrt{d_k}}\right)
},
\quad j\in\mathcal{N}(t),
\end{align}
where $\mathbf{q}_t$, $\mathbf{k}_j$, and $\mathbf{v}_j$ are the query, key, and value vectors, respectively, and $d_k$ is the dimension of the key-vector. The output representation is obtained by aggregating values over the same accessible set:
\begin{align}
\mathbf{o}_t
&=
\sum_{j\in\mathcal{N}(t)} \alpha_{tj}\mathbf{v}_j .
\end{align}
This formulation makes explicit that each decoding token can attend to all prefix tokens as persistent global context, while only attending locally within a bounded causal window over previously generated tokens. As a result, the model preserves access to the full prefix information while reducing the attention cost over the growing decode sequence.

\subsubsection{KV cache management}
For DeepSeek OCR baseline, it employs standard Multi-Head Attention (MHA)—the most classical form of attention, which offers strong expressiveness but imposes enormous KV cache pressure, the KV cache size is calculated as follows:
\begin{align}
C_{\mathrm{MHA}}(T) &= L_m + T.
\end{align}
In contrast, under R-SWA, the model always retains the full prefix cache of size $L_m$, but for the generated continuation it only needs to keep the most recent $n$ tokens. Therefore, after generating a total of $T$ tokens, the required KV cache size is
\begin{align}
C_{\mathrm{R\text{-}SWA}}(T) &= L_m + \min(n,\,T) \le L_m + n.
\end{align}
This shows that, unlike standard MHA whose cache size increases unboundedly with $T$, the decode-side cache of R-SWA is upper-bounded by a constant window size. To quantify the reduction, we define the cache ratio
\begin{align}
\rho(T)
&=
\frac{C_{\mathrm{R\text{-}SWA}}(T)}{C_{\mathrm{MHA}}(T)}
=
\frac{L_m+\min(n,\,T)}{L_m+T}.
\end{align}
If the generated length is sufficiently long such that $T\gg n$, then
\begin{align}
\rho(T)
&=
\frac{L_m+n}{L_m+T}.
\end{align}
which decreases as $T$ grows. In particular, when the decode length dominates both the prefix length and the window size, we have
\begin{align}
\rho(T)\approx \frac{L_m+n}{T}\to 0.
\end{align}
Therefore, for long-sequence decoding, R-SWA reduces the KV cache requirement from linear growth in $T$ to a bounded quantity $L_m+n$, yielding a substantial memory saving compared with standard MHA. Accordingly, R-SWA serves as the cornerstone to enabling near-unlimited parsing works under limited resources.

\subsubsection{Kernel study}

\begin{wrapfigure}{r}{0.5\textwidth}
\vspace{-5ex}
	\centering
	\includegraphics[width=0.5\textwidth]{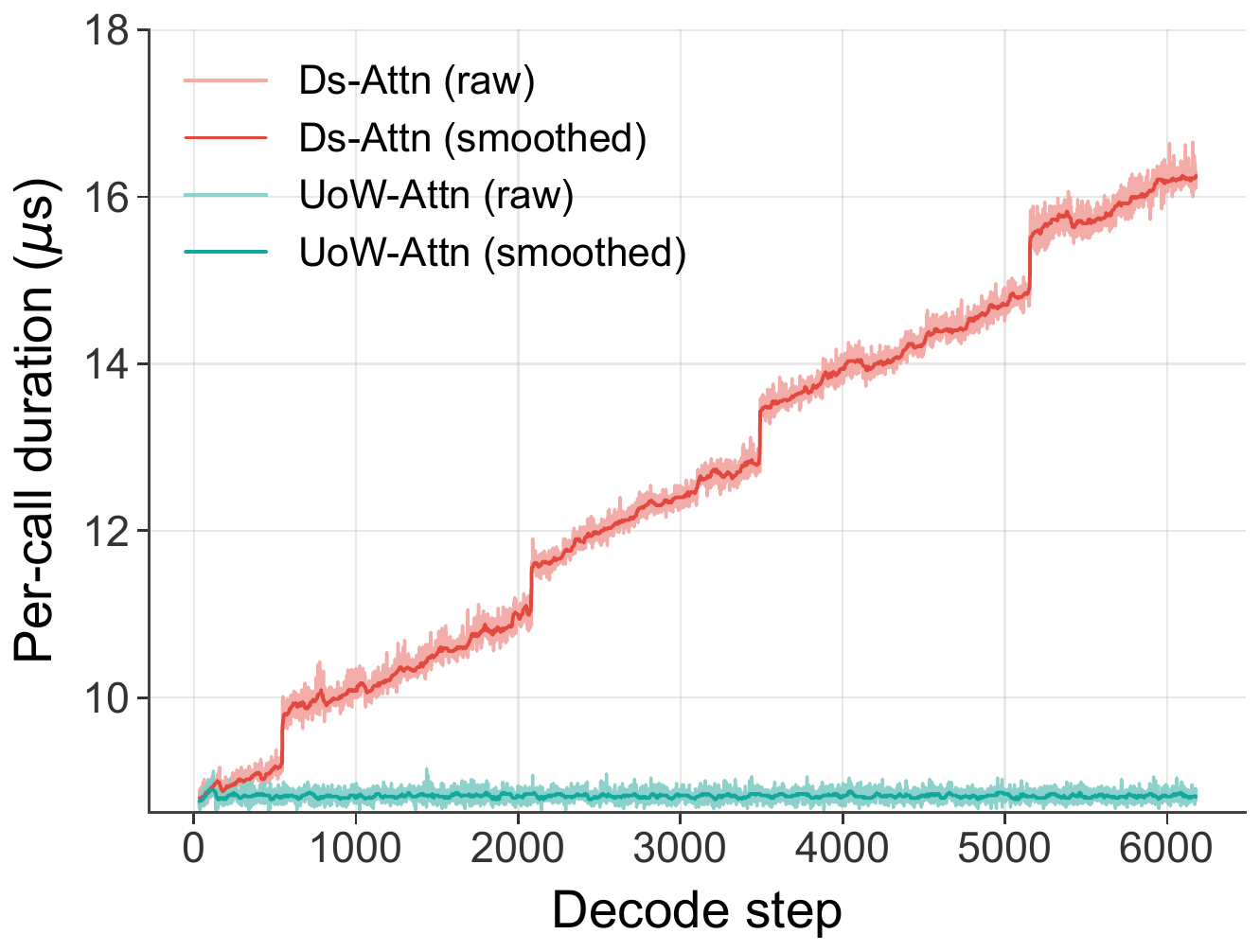}
	\caption{The latency of the Flash Attention v3 kernel as decoding length increases. }
	\label{fig3}
\vspace{-3ex}
\end{wrapfigure}


As shown in Figure~\ref{fig3}, we plot the per-call duration of the Flash Attention v3 kernel for both the DeepSeek OCR baseline and Unlimited OCR Works (denoted as UOW in the figure). The figure clearly shows that the standard MHA kernel in DeepSeek OCR incurs growing latency with each successive decoding step, whereas in Unlimited OCR the duration remains constant—a direct benefit of adopting R-SWA across all layers of the LLM decoder. The spike in the DeepSeek OCR occurs when the KV cache length crosses a certain alignment boundary, causing an abrupt drop in data transfer efficiency; this issue also does not arise with R-SWA. Besides, the same pattern will hold for GPU memory usage during inference: in the original DeepSeek OCR it scales linearly, while in Unlimited OCR it stays fixed. This joint stability in both computational cost and memory footprint is precisely what makes long-horizon parsing possible.

\section{Experimental Settings}

\subsection{Data Engine}
\label{data}

We construct approximately 2 million document OCR data samples to train Unlimited OCR, with a 9:1 ratio of single-page to multi-page data. For the single-page PDF data, we use Paddle OCR~\cite{cui2025paddleocr} for annotation, concatenating the coordinates and content of each block to construct end-to-end detection and parsing ground truth. The coordinates of each element are normalized to the range of 0–1000. All multi-page data are synthesized by concatenating single-page data. We randomly generate around 200k samples, each consisting of 2 to 50 pages, with \textit{<page>} used as a separator between pages.  All data are packed into a sequence length of 32K tokens.


\subsection{Implementation Details}

Starting from the DeepSeek OCR checkpoint~\cite{wei2025deepseek}, we continue training Unlimited OCR for 4,000 steps with a global batch size of 256 and a maximum sequence length of 32K on 8$\times$16 A800 GPUs, using random packing for all data. During training, we freeze the DeepEncoder and only train the LLM parameters, as the DeepEncoder is already sufficiently optimized in DeepSeek OCR. We use the AdamW~\cite{AdamW} optimizer and a cosine annealing scheduler~\cite{loshchilov2016sgdr} with an initial learning rate of 1e-4. To support 32K training, we adopt DeepEP~\cite{deepseek32}, with expert parallelism (EP) set to 4. The entire training pipeline is built on the Megatron-LM~\cite{megatron-lm} framework. For inference, we implement KV cache management for R-SWA in the Transformers library, along with corresponding support and optimizations in the SGLang inference engine. Both inference frameworks can operate Unlimited OCR under constant TPS (tokens/S) and GPU memory.

\section{Evaluation}

\subsection{Benchmark and Metrics}

We select OmniDocBench~\cite{ouyang2025omnidocbench} as the main benchmark for evaluating foundational document OCR capabilities, and test the Unlimited OCR on both v1.5 and v1.6 versions. OmniDocBench v1.6 includes 296 more test images than v1.5 and represents the latest benchmark, while v1.5 provides official metrics from more classic models—including our baseline DeepSeek OCR—which facilitates performance comparisons. For long-horizon OCR evaluation, an in-house test set is constructed, where we select a number of novels, documents, and papers and divide them by page count to assess the multi-page performance of Unlimited OCR. Specifically, we select books of 2, 5, 10, 20, and 40+ pages for testing, with no fewer than ten books for each category.

OmniDocBench is designed to evaluate document parsing capabilities across multiple dimensions, including text recognition, formula recognition, table structure extraction, and reading order prediction. It adopts task-specific metrics for a well-rounded evaluation: (1) Text Edit Distance (Edit $\downarrow$), which measures character-level accuracy for text recognition; (2) Formula CDM (CDM $\uparrow$), which evaluates the quality of mathematical formula recognition; (3) Table TEDS (TEDS $\uparrow$) and Table TEDS-S (TEDS-S $\uparrow$), which assess table structure extraction accuracy with and without content recognition; and (4) Reading Order Edit Distance (Edit $\downarrow$), which quantifies the correctness of predicted reading sequences. The overall score is then computed as a weighted average across text, formula, and table recognition tasks. For the in-house benchmark, we report both the Distinct-n and the Edit Distance. Distinct-n is the ratio of the number of unique n-grams to the total number of n-grams in the generated text.

\begin{table}[h]
\footnotesize
\caption{Comparison on OmniDocBench (v1.5/v1.6). All models in the table are end-to-end VLM-based architectures. v1.5 is primarily intended for comparison with classic end-to-end algorithms and the baseline DeepSeek OCR. v1.6 mainly compares against current end-to-end SOTA models. Except for the proposed Unlimited OCR, all other models are selected from the OmniDocBench repository.}
\label{tab:omnidocbench}
\centering
\setlength{\tabcolsep}{1.6pt}
{
\begin{tabular}{l|c|c|ccccc}
\toprule[.9pt]
{\textbf{Model}} &Size & Overall $\uparrow$ & Text$^{Edit}$ $\downarrow$  &  Formula$^{CDM}$ $\uparrow$ & Table$^{TEDs}$ $\uparrow$ & Table$^{TEDS_s}$ $\uparrow$ & Read-order$^{Edit}$ $\downarrow$  \\  
\midrule  

\multicolumn{8}{c}{\textbf{    End-to-end Model (v1.5)}} \\ 
\midrule 
OCRFlux~\cite{ocrflux} &3B &74.82 & 0.193 & 68.03 & 75.75 & 80.23 & 0.202 \\
GPT-4o~\cite{GPT4} & - &75.02 & 0.217 & 79.70 & 67.07 & 76.09 & 0.148 \\
InternVL3~\cite{zhu2025internvl3} & 78B & 80.33 & 0.131 & 83.42 & 70.64 & 77.74  & 0.113 \\
POINTS-Reader~\cite{liu2025pointsreader}& 3B & 80.98 & 0.134 & 79.20 & 77.13 & 81.66 & 0.145 \\
olmOCR~\cite{poznanski2025olmocr}& 7B   & 81.79 & 0.096 & 86.04 & 68.92 & 74.77 & 0.121 \\
InternVL3.5~\cite{wang2025internvl35}& 241B & 82.67 & 0.142 & 87.23 & 75.00 & 81.28  & 0.125 \\
MinerU2-VLM~\cite{wang2024mineru} &0.9B & 85.56 & 0.078 & 80.95 & 83.54 & 87.66  & 0.086 \\
Nanonets-OCR-s~\cite{NanonetsOCRs}&3B&85.59 & 0.093 & 85.90 & 80.14 & 85.57  & 0.108 \\
Qwen2.5-VL~\cite{Qwen2.5-VL} & 72B & 87.02 & 0.094 & 88.27 & 82.15 & 86.22 & 0.102 \\
Gemini-2.5 Pro\cite{google_gemini_web}&- & 88.03 & 0.075 & 85.82 & 85.71 & 90.29 & 0.097 \\
dots.ocr~\cite{dots}&3B & 88.41 & 0.048 & 83.22 & 86.78 & 90.62 & 0.053 \\ 
OCRVerse~\cite{OCRVerse} &4B & 88.56 & 0.058 & 86.91 & 84.55 & 88.45 & 0.071 \\ 
Qwen3-VL\cite{bai2025qwen3vltechnicalreport}&235B & 89.15 & 0.069 & 88.14 & 86.21 & 90.55 & 0.068 \\ 
DeepSeek-OCR 2~\cite{wei2026deepseek}&3B-A0.5B & 89.17  & 0.049  & 86.85  & 85.60  & 90.06 &  0.060 \\ 

\midrule 
DeepSeek-OCR &3B-A0.5B & 87.01 & 0.073 & 83.37 & 84.97 & 88.80 & 0.086 \\ 

\rowcolor{gray!10}
Unlimited-OCR &3B-A0.5B & 93.23 & 0.038 & 92.61 & 90.93 & 94.07 & 0.045 \\ 

 & & \textcolor{blue}{$\uparrow 6.22$} &\textcolor{blue}{$\downarrow 0.035$} & \textcolor{blue}{$\uparrow 9.24$} & \textcolor{blue}{$\uparrow 5.96$} & \textcolor{blue}{$\uparrow 5.27$} & \textcolor{blue}{$\downarrow 0.041$} \\

\midrule  
\multicolumn{8}{c}{\textbf{    End-to-end Model (v1.6)}} \\ 
\midrule  
HunyuanOCR~\cite{team2025hunyuanocr}&1B & 89.95  & 0.088  & 87.68  & 91.01  & 92.23 &  0.171 \\ 
DeepSeek-OCR 2~\cite{wei2026deepseek}&3B-A0.5B & 90.25  & 0.050  & 91.84  & 83.89  & 87.75 &  0.144 \\ 
dots.ocr~\cite{dots}&3B & 90.77 & 0.048 & 89.95 & 87.18 & 90.58 & 0.138 \\ 
FireRed-OCR~\cite{wu2026firered}&2B & 93.26 & 0.037  & 95.44  & 88.04  & 91.06 &  0.131 \\ 
Logics-Parsing-v2~\cite{Logics-Parsing-V2}&4B & 93.33  & 0.041  & 95.65  & 88.42  & 91.98 &  0.137 \\ 
Qianfan-OCR~\cite{dong2026qianfan}&4B & 93.90  & 0.040  & 95.08  & 90.53  & 93.31 &  0.13 \\ 
\midrule 
\rowcolor{gray!10}
Unlimited-OCR &3B-A0.5B & 93.92 & 0.042 & 95.79 & 90.16 & 93.32 & 0.129 \\ 
\bottomrule[.9pt]
\end{tabular}
}
\end{table}

\subsection{Main Results}
\label{OmniDocbench v1.5}

As shown in Table~\ref{tab:omnidocbench}, by continue-training on merely 2M PDF-document-specific data based on DeepSeek OCR, Unlimited OCR achieves end-to-end SOTA performance. This demonstrates the effectiveness of R-SWA on parsing tasks. First, compared with the standard attention in DeepSeek OCR, R-SWA may allow the model to focus more on dense OCR tasks, whereas full attention could lead to divergence as the output length increases. On the other hand, the state transition across intra-page content under R-SWA is both workable and solid. Specifically, on OmniDocBench v1.5, compared with DeepSeek OCR, the text edit distance drops by 0.035, and the table TEDS improves by 5.96\%, indicating that historical information is causally and continuously fed into the sliding window, enabling the model to clearly locate its OCR progress even though it sees only a few tokens. On the OmniDocBench v1.6 benchmark, Unlimited OCR again achieves end-to-end SOTA (93.92\% on overall metric), further proving that for single-page PDF-level document OCR tasks, replacing all standard attention entirely with R-SWA of width 128 is both effective and lossless.

Moreover, Unlimited OCR gains all the benefits of DeepSeek OCR, such as the MoE architecture with only 0.5B activated parameters, resulting in very high inference efficiency. In the OmniDocBench, Unlimited OCR achieves 5580 TPS (tokens/s/512 concurrency) compared to DeepSeek OCR’s 4951 TPS under ''Base" DeepEncoder mode, representing a 12.7\% speed increase. Of course, the average document length in OmniDocBench is relatively short—the longer the output length, the more pronounced the advantage of Unlimited OCR becomes.

        
		



\subsection{Subcategory Study}

OmniDocBench (v1.5) provides 9 types of documents, and conducting a subcategory comparison is crucial for a more systematic and comprehensive analysis of R-SWA. As shown in Table~\ref{table-3}, compared to DeepSeek OCR, Unlimited OCR shows clear and consistent gains across every metric, demonstrating that our decoder-side optimization, \textit{i.e.}, R-SWA, delivers a genuine "free lunch"—improvements without compromises. Compared to DeepSeek OCR 2, Unlimited OCR also holds a clear advantage, with seven-ninths of both the text edit distance and reading order scores surpassing those of DeepSeek OCR 2. For documents with complex layouts such as PPT, newspapers, magazines, and note, Unlimited OCR shows no disadvantage either, further demonstrating that replacing all standard attention with R-SWA for LLM-decoder is complete and sound for parsing tasks.

\begin{table}[ht]\small
    \centering	

    \caption{Detailed subcategory comparison between Unlimited OCR and the DeepSeek-OCR series across nine document types. R-order denotes reading order. All metrics are edit distances, where lower is better. Red cells indicate that the corresponding metric of DeepSeek-OCR or DeepSeek-OCR 2 is better than that of Unlimited OCR.}
    \setlength{\abovecaptionskip}{0.2cm}
    \setlength{\tabcolsep}{0.6mm}{

        \begin{tabular}{l|c|ccccccccc}
                \toprule
            Model & Edit $\downarrow$ & PPT & \makecell{Academic \\ Paper} & Book & \makecell{Colorful \\ Textbook} & \makecell{Exam \\ Paper} & Magazine & Newspaper & Note & \makecell{Research \\ Report}  \\
                \midrule
                \multirow{2}{*}{DS-OCR} & Text & 0.052 & 0.028 & 0.022 & 0.130 & 0.074 & 0.049 & 0.131 & 0.145 & 0.015   \\
                  & R-order & 0.052 & 0.021 & 0.040 & 0.125 & 0.083 & 0.101 & 0.217 & 0.089 & 0.016   \\
                \midrule
                 \multirow{2}{*}{DS-OCR 2}  &Text & 0.031 & \cellcolor{pink!40}{0.013} & 0.033 & 0.053 & \cellcolor{pink!40}{0.047} & 0.026 & 0.139 & 0.068 & 0.008  \\
                         &R-order & 0.025 & 0.013 & 0.027 & 0.066 & \cellcolor{pink!40}{0.048} & 0.100 & 0.176 & 0.035 & \cellcolor{pink!40}{0.011}   \\

                                \midrule
                 \multirow{2}{*}{UOW}  &Text & 0.025 & 0.023 & 0.019 & 0.046 & 0.049 & 0.020 & 0.081 & 0.066 & 0.008  \\
                         &R-order & 0.023 & 0.012 & 0.025 & 0.051 & 0.049 & 0.061 & 0.134 & 0.018 & 0.013   \\
                
    \bottomrule
    \end{tabular}}

    \label{table-3}
	
\end{table}

\begin{table}[!t]\small
    \centering

    \caption{Performance of long-horizon OCR. We test the distinct-n and edit distance under different page numbers. Distinct-n is the higher the better.}
    \setlength{\abovecaptionskip}{0.5cm}
    \setlength{\tabcolsep}{3.5mm}{

        \begin{tabular}{l|cccccc}
                \toprule
            \diagbox[height=0.61cm]{Metric}{Pages} & 2 & 5 & 10 &  15 & 20 & 40+  \\
                \midrule
                Distinct-20 $\uparrow$ & 99.76\% & 99.78\% & 97.49\% & 99.92\%  & 98.73\%  & 96.08\%  \\
                Distinct-35 $\uparrow$ & 99.87\% & 99.98\% & 99.83\% & 99.99\%  & 99.89\%  & 96.90\%  \\

                \midrule
                 Edit Distance $\downarrow$ & 0.0362 & 0.0452 & 0.0526  & 0.0787 & 0.0572 & 0.1069   \\

    \bottomrule
    \end{tabular}}


	\label{table4}
    \end{table}

\subsection{Long-horizon Parsing}

Long-horizon parsing is one of the novel capabilities of Unlimited OCR. Two main obstacles have hindered previous models from achieving this: first, excessively long output sequences can easily exceed the maximum token limit; second, output latency grows with sequence length, causing the OCR of documents spanning dozens of pages to become progressively slower. Unlimited OCR, equipped with R-SWA, can prefill tens to hundreds of document pages in a single pass and parse continuously from the first page to the last. Throughout this process, the KV cache remains fixed, so output latency stays constant—making long-horizon parsing feasible. As shown in Table~\ref{table4}, our model delivers satisfactory performance in multi-page one-shot OCR scenarios, maintaining strong results even with 20 pages input simultaneously. At 40+ pages, the edit distance remains below 0.11 along with 97\% Distinct-35. We examine the cases with repeated errors and find that most occur where small text in the PDF is difficult to discern, primarily due to the use of DeepEncoder's "Base" mode (1024$\times$1024 resolution) under multi-page conditions, rather than R-SWA losing direction in long-horizon parsing process. 




\begin{table}[!t]\small
    \centering

    \caption{Theoretical inference performance ceiling comparison. We compare the TPS upper limits of DeepSeek OCR and Unlimited OCR across varying output lengths.}
    \setlength{\abovecaptionskip}{0.5cm}
    \setlength{\tabcolsep}{2.5mm}{

        \begin{tabular}{l|ccccccc}
                \toprule
            \diagbox[height=0.7cm]{Model}{TPS} & 256 & 512 & 1024 &  2048 & 3072 & 4096 & 6144  \\
                \midrule
                Deepseek OCR  & 7229.32 & 7468.27 & 7422.50 & 7166.85  & 6790.72  & 6430.21 & 5822.87\\
                Unlimited OCR  & 7229.52 & 7714.78 & 7840.94& 7881.11  & 7881.93  & 7905.18  & 7847.71 \\

    \bottomrule
    \end{tabular}}


	\label{table5}
    \end{table}

\section{Efficiency Analysis}
As presented in Table~\ref{table5}, we compare the output tokens per second (TPS) of Unlimited OCR and DeepSeek OCR under ideal concurrency conditions. The prefill length is fixed at 10, with all other settings held identical. The results show that at 256 tokens, the inference speeds of the two models are virtually the same. As the output length grows, however, the TPS of DeepSeek OCR steadily declines, and at 6,000 tokens, it lags behind Unlimited OCR—which incorporates R-SWA—by 35\%. These findings further validate the effectiveness of R-SWA and underscore that consistent generation speed is a critical requirement for long-horizon OCR tasks.

\section{Limitation and Future Work}

Our model cannot achieve truly unlimited parsing under a finite context length (\textit{e.g.,} 32K), as it is also constrained by the prefill length. Although DeepEncoder already achieves a high compression rate for image tokens, the prefill still becomes very long as the number of pages accumulates. In the short term, we will train models with longer context lengths, such as 128K, to support the prefill of more pages. In the long term, we plan to build a prefill pool and enable the model to learn to automatically fetch prefill KV chunks, thereby simulating the effect of a human flipping through pages, so as to achieve truly unlimited OCR works. In addition, we will also transfer R-SWA to reference-based tasks such as ASR and translation.

\section{Conclusion}
In this technical report, we propose the Unlimited OCR model and present the R-SWA algorithm to support its capability for long-horizon parsing. We verify that when all standard attention in the decoder of an end-to-end model is replaced with causal reference-based SWA, the model's performance on parsing tasks remains lossless. This indicates that the model learns to continuously pass useful information from historical outputs into the window, and this soft form of forgetting is consistent with how we humans behave when transcribing a book. We believe that R-SWA will be applied to more tasks in the future, making attention computation and memory footprint no longer the bottleneck for long-horizon parsing field.

\section{Author List}

\noindent * indicates project leader; $^\dag$ indicates technical director

\noindent \textbf{Core Contributors:} Youyang Yin, Huanhuan Liu*, YY$^\dag$

\noindent \textbf{Contributors:} Qunyi Xie, Chaorun Liu, Shiqi Yang, Shaohua Wang, Zhanlong Liu, Hao Zou, Jinyue Chen, Shu Wei, Jingjing Wu, Mingxin Huang, Zhen Wu, Guibin Wang, Tengyu Du,  Lei Jia


\bibliography{main}

@article{wei2025deepseek,
  title={Deepseek-ocr: Contexts optical compression},
  author={Wei, Haoran and Sun, Yaofeng and Li, Yukun},
  journal={arXiv preprint arXiv:2510.18234},
  year={2025}
}

@article{wei2026deepseek,
  title={DeepSeek-OCR 2: Visual Causal Flow},
  author={Wei, Haoran and Sun, Yaofeng and Li, Yukun},
  journal={arXiv preprint arXiv:2601.20552},
  year={2026}
}

@article{wu2026firered,
  title={Firered-ocr technical report},
  author={Wu, Hao and Lou, Haoran and Li, Xinyue and Zhong, Zuodong and Sun, Zhaojun and Chen, Phellon and Zhou, Xuanhe and Zuo, Kai and Chen, Yibo and Tang, Xu and others},
  journal={arXiv preprint arXiv:2603.01840},
  year={2026}
}

@article{megatron-lm,
  title={Megatron-LM: Training Multi-Billion Parameter Language Models Using Model Parallelism},
  author={Shoeybi, Mohammad and Patwary, Mostofa and Puri, Raul and LeGresley, Patrick and Casper, Jared and Catanzaro, Bryan},
  journal={arXiv preprint arXiv:1909.08053},
  year={2019}
}

@inproceedings{liu2025pointsreader,
  author    = {Yuan Liu and Z. Zhao and L. Tian and others},
  title     = {POINTS-Reader: Distillation-Free Adaptation of Vision-Language Models for Document Conversion},
  booktitle = {Proceedings of the 2025 Conference on Empirical Methods in Natural Language Processing},
  year      = {2025},
  pages     = {1576--1601},
  month     = {November}
}

@article{wang2025internvl35,
  author    = {Weiyun Wang and Z. Gao and L. Gu and others},
  title     = {InternVL3.5: Advancing Open-Source Multimodal Models in Versatility, Reasoning, and Efficiency},
  journal   = {arXiv preprint arXiv:2508.18265},
  year      = {2025}
}

@article{team2023gemini,
  title={Gemini: a family of highly capable multimodal models},
  author={Team, Gemini and Anil, Rohan and Borgeaud, Sebastian and Alayrac, Jean-Baptiste and Yu, Jiahui and Soricut, Radu and Schalkwyk, Johan and Dai, Andrew M and Hauth, Anja and Millican, Katie and others},
  journal={arXiv preprint arXiv:2312.11805},
  year={2023}
}

@article{bai2025qwen3vltechnicalreport,
  author    = {S. Bai and Y. Cai and R. Chen and others},
  title     = {Qwen3-VL Technical Report},
  journal   = {arXiv preprint arXiv:2511.21631},
  year      = {2025},
  url       = {https://arxiv.org/abs/2511.21631}
}

@inproceedings{li2023blip,
  title={Blip-2: Bootstrapping language-image pre-training with frozen image encoders and large language models},
  author={Li, Junnan and Li, Dongxu and Savarese, Silvio and Hoi, Steven},
  booktitle={International conference on machine learning},
  pages={19730--19742},
  year={2023},
  organization={PMLR}
}

@inproceedings{ouyang2025omnidocbench,
  title={Omnidocbench: Benchmarking diverse pdf document parsing with comprehensive annotations},
  author={Ouyang, Linke and Qu, Yuan and Zhou, Hongbin and Zhu, Jiawei and Zhang, Rui and Lin, Qunshu and Wang, Bin and Zhao, Zhiyuan and Jiang, Man and Zhao, Xiaomeng and others},
  booktitle={Proceedings of the Computer Vision and Pattern Recognition Conference},
  pages={24838--24848},
  year={2025}
}

@inproceedings{wei2024vary,
  title={Vary: Scaling up the vision vocabulary for large vision-language model},
  author={Wei, Haoran and Kong, Lingyu and Chen, Jinyue and Zhao, Liang and Ge, Zheng and Yang, Jinrong and Sun, Jianjian and Han, Chunrui and Zhang, Xiangyu},
  booktitle={European Conference on Computer Vision},
  pages={408--424},
  year={2024},
  organization={Springer}
}

@article{Qwen-VL,
  title={Qwen-VL: A Versatile Vision-Language Model for Understanding, Localization, Text Reading, and Beyond},
  author={Bai, Jinze and Bai, Shuai and Yang, Shusheng and Wang, Shijie and Tan, Sinan and Wang, Peng and Lin, Junyang and Zhou, Chang and Zhou, Jingren},
  journal={arXiv preprint arXiv:2308.12966},
  year={2023}
}

@article{kirillov2023segment,
  title={Segment anything},
  author={Kirillov, Alexander and Mintun, Eric and Ravi, Nikhila and Mao, Hanzi and Rolland, Chloe and Gustafson, Laura and Xiao, Tete and Whitehead, Spencer and Berg, Alexander C and Lo, Wan-Yen and others},
  journal={arXiv preprint arXiv:2304.02643},
  year={2023}
}

@inproceedings{radford2021learning,
  title={Learning transferable visual models from natural language supervision},
  author={Radford, Alec and Kim, Jong Wook and Hallacy, Chris and Ramesh, Aditya and Goh, Gabriel and Agarwal, Sandhini and Sastry, Girish and Askell, Amanda and Mishkin, Pamela and Clark, Jack and others},
  booktitle={International conference on machine learning},
  pages={8748--8763},
  year={2021},
  organization={PMLR}
}

@article{blecher2023nougat,
  title={Nougat: Neural optical understanding for academic documents},
  author={Blecher, Lukas and Cucurull, Guillem and Scialom, Thomas and Stojnic, Robert},
  journal={arXiv preprint arXiv:2308.13418},
  year={2023}
}

@article{wei2024general,
  title={General ocr theory: Towards ocr-2.0 via a unified end-to-end model},
  author={Wei, Haoran and Liu, Chenglong and Chen, Jinyue and Wang, Jia and Kong, Lingyu and Xu, Yanming and Ge, Zheng and Zhao, Liang and Sun, Jianjian and Peng, Yuang and others},
  journal={arXiv preprint arXiv:2409.01704},
  year={2024}
}

@article{deepseek32,
  title={Deepseek-v3. 2: Pushing the frontier of open large language models},
  author={Liu, Aixin and Mei, Aoxue and Lin, Bangcai and Xue, Bing and Wang, Bingxuan and Xu, Bingzheng and Wu, Bochao and Zhang, Bowei and Lin, Chaofan and Dong, Chen and others},
  journal={arXiv preprint arXiv:2512.02556},
  year={2025}
}

@article{huang2026step3,
  title={STEP3-VL-10B Technical Report},
  author={Huang, Ailin and Yao, Chengyuan and Han, Chunrui and Wan, Fanqi and Guo, Hangyu and Lv, Haoran and Zhou, Hongyu and Wang, Jia and Zhou, Jian and Sun, Jianjian and others},
  journal={arXiv preprint arXiv:2601.09668},
  year={2026}
}

@article{wang2024mineru,
  title={Mineru: An open-source solution for precise document content extraction},
  author={Wang, Bin and Xu, Chao and Zhao, Xiaomeng and Ouyang, Linke and Wu, Fan and Zhao, Zhiyuan and Xu, Rui and Liu, Kaiwen and Qu, Yuan and Shang, Fukai and others},
  journal={arXiv preprint arXiv:2409.18839},
  year={2024}
}

@article{poznanski2025olmocr,
  title={olmocr: Unlocking trillions of tokens in pdfs with vision language models},
  author={Poznanski, Jake and Rangapur, Aman and Borchardt, Jon and Dunkelberger, Jason and Huff, Regan and Lin, Daniel and Wilhelm, Christopher and Lo, Kyle and Soldaini, Luca},
  journal={arXiv preprint arXiv:2502.18443},
  year={2025}
}

@article{cui2025paddleocr,
  title={Paddleocr 3.0 technical report},
  author={Cui, Cheng and Sun, Ting and Lin, Manhui and Gao, Tingquan and Zhang, Yubo and Liu, Jiaxuan and Wang, Xueqing and Zhang, Zelun and Zhou, Changda and Liu, Hongen and others},
  journal={arXiv preprint arXiv:2507.05595},
  year={2025}
}

@inproceedings{AdamW,
  author       = {Ilya Loshchilov and
                  Frank Hutter},
  title        = {Decoupled Weight Decay Regularization},
  booktitle    = {{ICLR}},
  year         = {2019}
}

@article{loshchilov2016sgdr,
  title={Sgdr: Stochastic gradient descent with warm restarts},
  author={Loshchilov, Ilya and Hutter, Frank},
  journal={arXiv preprint arXiv:1608.03983},
  year={2016}
}

@article{team2025hunyuanocr,
  title={Hunyuanocr technical report},
  author={Team, Hunyuan Vision and Lyu, Pengyuan and Wan, Xingyu and Li, Gengluo and Peng, Shangpin and Wang, Weinong and Wu, Liang and Shen, Huawen and Zhou, Yu and Tang, Canhui and others},
  journal={arXiv preprint arXiv:2511.19575},
  year={2025}
}

@article{dong2026qianfan,
  title={Qianfan-ocr: A unified end-to-end model for document intelligence},
  author={Dong, Daxiang and Zheng, Mingming and Xu, Dong and Luo, Chunhua and Zhuang, Bairong and Li, Yuxuan and He, Ruoyun and Wang, Haoran and Zhang, Wenyu and Wang, Wenbo and others},
  journal={arXiv preprint arXiv:2603.13398},
  year={2026}
}

@misc{Logics-Parsing-V2,
  author = {alibaba},
  year = {2026},
  url = {https://github.com/alibaba/Logics-Parsing},
}

@article{feng2025dolphin,
  title={Dolphin: Document image parsing via heterogeneous anchor prompting},
  author={Feng, Hao and Wei, Shu and Fei, Xiang and Shi, Wei and Han, Yingdong and Liao, Lei and Lu, Jinghui and Wu, Binghong and Liu, Qi and Lin, Chunhui and others},
  journal={arXiv preprint arXiv:2505.14059},
  year={2025}
}

@article{li2025monkeyocr,
  title={MonkeyOCR: Document Parsing with a Structure-Recognition-Relation Triplet Paradigm},
  author={Li, Zhang and Liu, Yuliang and Liu, Qiang and Ma, Zhiyin and Zhang, Ziyang and Zhang, Shuo and Guo, Zidun and Zhang, Jiarui and Wang, Xinyu and Bai, Xiang},
  journal={arXiv preprint arXiv:2506.05218},
  year={2025}
}

@article{Qwen2.5-VL,
  title={Qwen2.5-VL Technical Report},
  author={Bai, Shuai and Chen, Keqin and Liu, Xuejing and Wang, Jialin and Ge, Wenbin and Song, Sibo and Dang, Kai and Wang, Peng and Wang, Shijie and Tang, Jun and Zhong, Humen and Zhu, Yuanzhi and Yang, Mingkun and Li, Zhaohai and Wan, Jianqiang and Wang, Pengfei and Ding, Wei and Fu, Zheren and Xu, Yiheng and Ye, Jiabo and Zhang, Xi and Xie, Tianbao and Cheng, Zesen and Zhang, Hang and Yang, Zhibo and Xu, Haiyang and Lin, Junyang},
  journal={arXiv preprint arXiv:2502.13923},
  year={2025}
}

@article{cui2025paddleocrvl,
  title={Paddleocr-vl: Boosting multilingual document parsing via a 0.9 b ultra-compact vision-language model},
  author={Cui, Cheng and Sun, T and Liang, S and others},
  journal={arXiv preprint arXiv:2510.14528},
  year={2025}
}

@misc{ocrflux,
  author = {},
  title = {OCRFlux},
  year = {2025},
  url = {https://github.com/chatdoc-com/OCRFlux},
}

@misc{OCRVerse,
  author = {},
  title = {OCRVerse},
  year = {2025},
  url = {https://github.com/DocTron-hub/OCRVerse},
}

@misc{NanonetsOCRs,
  author = {},
  title = {Nanonets-OCR-s},
  year = {2025},
  url = {https://huggingface.co/nanonets/Nanonets-OCR-s},
}

@misc{GPT4,
      title={GPT-4 Technical Report}, 
      author={OpenAI},
      year={2023},
      eprint={arXiv preprint arXiv:2303.08774}
}

@article{zhu2025internvl3,
  title={Internvl3: Exploring advanced training and test-time recipes for open-source multimodal models},
  author={Zhu, Jinguo and Wang, Weiyun and Chen, Zhe and Liu, Zhaoyang and Ye, Shenglong and Gu, Lixin and Tian, Hao and Duan, Yuchen and Su, Weijie and Shao, Jie and others},
  journal={arXiv preprint arXiv:2504.10479},
  year={2025}
}

@misc{dots,
      title={dots.ocr}, 
      author={Rednote},
      year={2025},
      url = {https://github.com/rednote-hilab/dots.ocr},
}

@misc{google_gemini_web,
  author = {Google AI},
  title = {Gemini 2.5-Pro},
  year = {2025},
  url = {https://gemini.google.com/},
}

\end{CJK*}
\end{document}